\documentclass[preprint,12pt,authoryear]{elsarticle}
\usepackage{orcidlink}
\usepackage{tabularx}   
\usepackage{longtable}
\usepackage{makecell}

\newlength{\RiskW}
\newcolumntype{Y}{>{\raggedright\arraybackslash}X} 
\newcolumntype{C}[1]{>{\centering\arraybackslash}m{#1}}
\usepackage{multicol}
\usepackage{multirow}
\usepackage{array}
\usepackage{amssymb}
\usepackage{amsmath}
\usepackage{graphicx} 
\usepackage{algorithm} 
\usepackage{algorithmic} 
\usepackage{booktabs}
\usepackage{url}  
\usepackage{xcolor} 
\usepackage{float} 
\usepackage{caption} 
\usepackage{subcaption} 
\usepackage{hyperref} 
\usepackage{bm} 
\usepackage{newtxtext,newtxmath} 
\usepackage{helvet}  
\usepackage{courier} 
\usepackage{type1cm}  

\usepackage{amssymb}
\usepackage{amsmath}

\journal{Nuclear Physics B}

\hypersetup{
    colorlinks=true,
    linkcolor=blue,
    filecolor=magenta,      
    urlcolor=cyan,
    citecolor=blue,
    pdftitle={Your Paper Title},
    pdfauthor={Your Name}
}

\floatname{algorithm}{Algorithm}

\journal{Nuclear Physics B}

\begin{document}

\begin{frontmatter}



\title{MARS: Multi-Agent Robotic System with Multimodal Large Language Models for Assistive Intelligence} 

\author[inst1]{Renjun Gao\,\orcidlink{0009-0005-0028-428X}\corref{cor1}}
\ead{anakbhs@gmail.com}



\affiliation[inst1]{%
  organization={School of Computer Science and Engineering, Faculty of Innovation Engineering, Macau University of Science and Technology},%
  city={Macao Special Administrative Region of China}
}

\cortext[cor1]{Corresponding author.}

\begin{abstract}
Multimodal large language models (MLLMs) have shown remarkable capabilities in cross-modal understanding and reasoning, offering new opportunities for intelligent assistive systems, yet existing systems still struggle with risk-aware planning, user personalization, and grounding language plans into executable skills in cluttered homes. We introduce \textbf{MARS}, a \textbf{M}ulti-\textbf{A}gent \textbf{R}obotic \textbf{S}ystem powered by MLLMs for assistive intelligence and designed for smart home robots supporting people with disabilities. The system integrates four agents: a visual perception agent for extracting semantic and spatial features from environment images, a risk assessment agent for identifying and prioritizing hazards, a planning agent for generating executable action sequences, and an evaluation agent for iterative optimization. By combining multimodal perception with hierarchical multi-agent decision-making, the framework enables adaptive, risk-aware, and personalized assistance in dynamic indoor environments. Experiments on multiple datasets demonstrate the superior overall performance of the proposed system in risk-aware planning and coordinated multi-agent execution compared with state-of-the-art multimodal models. The proposed approach also highlights the potential of collaborative AI for practical assistive scenarios and provides a generalizable methodology for deploying MLLM-enabled multi-agent systems in real-world environments.
\end{abstract}




\end{frontmatter}



\section{Introduction}
In recent years, large language models (LLMs) based on the Transformer architecture have achieved breakthrough progress. Leveraging massive parameter scales and extensive general knowledge, LLMs have demonstrated near-human intelligence in natural language processing (NLP) \cite{nasution2024chatgpt} and content generation tasks \cite{agossah2023llm}. However, unimodal language models are insufficient for addressing the complex perceptual and interactive demands of real-world scenarios \cite{janssens2024multi}, which has driven the rapid rise of multimodal large language models (MLLMs). By integrating visual and linguistic information, MLLMs enable cross-modal semantic understanding and knowledge transfer, allowing them to handle tasks that are closer to daily-life complexity, such as image captioning \cite{bucciarelli2024personalizing}, visual question answering \cite{kuang2025natural,fang2025guided}, and scene reasoning \cite{zhao2025sce2drivex}.

A particularly important application scenario of MLLMs is indoor assistive agents for the elderly \cite{song2024elderease}. With an aging society and the increasing demand for independent living among people with disabilities, intelligent environmental assistance has become a critical research direction. The multimodal nature of environmental perception and the semantic complexity of risk assessment impose higher requirements on cross-modal understanding and decision-making capabilities. In this context, MLLMs, with their strong multimodal fusion and reasoning capabilities, provide a novel technical foundation for the construction of intelligent, context-aware assistive systems, serving as a core bridge between perception and decision-making \cite{cui2024survey,huang2023language}.

Existing advances have demonstrated preliminary potential in environmental perception and assistive applications. Some studies have employed cross-modal models such as CLIP and BLIP to build vision-language understanding frameworks, achieving basic functionalities such as indoor obstacle detection \cite{zhang2025vlm} and scene semantic annotation \cite{chen2023clip2scene}. Moreover, Huang et al. attempted to integrate LLMs with robotic control, generating simple action sequences from natural language instructions \cite{huang2023instruct2act}. However, current systems often focus on single-task modules, lacking dynamic evaluation of environmental risks and prioritization capabilities. Furthermore, there are significant limitations in multi-agent collaboration design: (i) reliance on fixed rule engines, and (ii) difficulty in resolving multi-objective conflicts in complex scenarios. These limitations prevent existing technologies from fully satisfying real-time, safe, and personalized assistance needs for people with disabilities in everyday environments.

For accurate assistive navigation, current research faces several challenges:(i) Precise perception and dynamic risk assessment in complex environments: Indoor scenes contain diverse obstacles, dynamically changing layouts, and varying user states, making it difficult for existing perception models to comprehensively extract risk features and perform dynamic prioritization.
(ii) Efficiency and robustness bottlenecks in multi-agent collaboration: Perception, reasoning, and planning modules are highly coupled, and existing collaboration models lack flexible task allocation and conflict resolution mechanisms, limiting their ability to handle dynamic multi-objective adjustments.
(iii) Lack of dedicated evaluation datasets: Most current datasets focus on a single modality or simple scenarios, lacking comprehensive benchmarks that annotate environmental obstacles, user states, and optimal actions, which constrains system training and performance validation.

To address these challenges, we propose a MLLM-based multi-agent collaborative system. For the assistive robotics domain, the input branch first preprocesses static images of the user’s surrounding environment and, by integrating depth and semantic features, extracts key objects/targets such as indoor layouts, object positions, and human presence. Based on this, a visual agent is constructed using an MLLM combined with models like CLIP, capable of recognizing and extracting environmental information such as obstacle types, locations, and passage widths, thereby satisfying perception requirements. Leveraging the perception output, the risk reasoning branch employs multi-agent collaboration mechanisms to analyze environmental information, assess potential hazards, and generate prioritized handling sequences. The planning and iterative branch then generates preliminary action sequences based on risk priority, which are evaluated by the evaluation agent to detect potential conflicts and risks, with feedback guiding subsequent adjustments. The system ultimately outputs robot-executable physical actions and path plans. Overall, the contributions of this work are as follows:

\begin{itemize}
    \item We target behavior decision and planning for smart home robots assisting people with disabilities, addressing dynamic environments and diverse obstacles to ensure safe and personalized assistance.
    \item We propose a multi-agent collaborative system based on MLLMs, integrating perception, reasoning, planning, and evaluation agents for cross-modal understanding and risk-aware decision-making.
    \item We evaluate the system on multiple datasets and compare it with state-of-the-art multimodal models, demonstrating improved perception, planning, and coordination performance in realistic assistive scenarios.
\end{itemize}

\section{Related Work}
\subsection{Multi-Agent Systems}
A multi-agent system (MAS) is a computerized system composed of multiple interacting agents, with key components including agents, the environment, communication mechanisms, and organizational or control structures. Agents are the core participants endowed with roles, capabilities, behaviors, and knowledge models, providing intelligence to themselves and the overall system through functionalities such as learning, planning, reasoning, and decision-making. The environment is the external world in which agents operate, perceive, and act, and it can be either simulated or physical, such as factories \cite{lim2024large}, road networks \cite{zhang2025ccma}, or power grids \cite{yao2025ai}. Communication among agents is typically conducted using standard agent communication languages, while organizational structures can vary from hierarchical architectures \cite{zou2024cooperative} to adaptive arrangements \cite{de2023emergent} tailored to practical needs.

The inherent learning and autonomous decision-making capabilities of individual agents provide MAS with high flexibility and adaptability. For example, Zhang et al. \cite{zhang2025webpilot} developed WebPilot, a multi-agent system with a dual optimization strategy that leverages an improved Monte Carlo tree search to handle complex web environments more effectively, achieving state-of-the-art performance on the WebArena dataset. Yu et al. \cite{yu2024fincon} designed FinCon, a LLM-based MAS for the financial domain, employing a “manager-analyst” hierarchical architecture to achieve high-quality decision-making while reducing unnecessary point-to-point communication costs. Aparna et al. \cite{kumari2024multi} proposed a decentralized real-time energy management (REM) method for modern residential power consumption, constructing a deep reinforcement learning-based ensemble model to enable efficient real-time allocation in the power grid. AgentCoord \cite{pan2025agentcoord} addresses coordination requirements in MAS by developing an LLM-based visual exploration framework; through a three-stage generative process, it converts users’ overall goals into initial strategies for visual organization, with user studies confirming its feasibility and effectiveness in agent collaboration processes.

\subsection{Collaborative Artificial Intelligence}
Collaborative artificial intelligence (Collaborative AI) refers to AI systems designed to work synergistically with other AI agents or humans \cite{przegalinska2025collaborative}. Its development is driven by two main factors: (I) the growing demand for AI systems capable of cooperating with other agents due to improvements in AI optimization and tool use, and (II) the efficiency gains achieved through positive collaboration among AI models. Research in this area spans multiple domains, including MAS \cite{selvam2024multi}, human-computer interaction \cite{huang2023human}, game theory \cite{xing2024deep}, and natural language processing \cite{babaian2023nlp}. By integrating these technologies, collaborative AI holds the potential to enable novel applications with significant economic and societal impact \cite{akinnagbe2024human}.

For instance, Wang et al. \cite{wang2024mobile} developed a smart assistant for mobile device navigation based on multi-agent collaboration, demonstrating notable advantages in stability and robustness. Islam et al. \cite{islam2024mapcoder} proposed a multi-agent framework based on cooperative-competitive mechanisms, incorporating fine-grained designs for problem description and multi-step complex reasoning, achieving results beyond existing solutions. Ghafarollahi et al. \cite{ghafarollahi2025sciagents} constructed a biomimetic MAS for graph reasoning architectures, enabling automated support for scientific discovery and offering feasible guidance for future research. Fourney et al. \cite{fourney2024magentic} developed Magentic-One, a general-purpose large model leveraging MAS for complex task-solving; experiments show that it achieves state-of-the-art performance across multiple domains.

As a technology focusing on coordination between humans and intelligent agents, collaborative AI exemplifies the practical extension of MAS principles to scenarios requiring dynamic, goal-driven cooperation. From stability optimization in mobile navigation and breakthroughs in complex reasoning mechanisms, to automation in scientific discovery and achieving SOTA results in complex tasks, these studies collectively demonstrate the strong potential of collaborative AI across diverse applications. Its continued development is expected to bring profound economic and societal benefits.

\section{System Design}
We propose a multi-agent collaboration framework built on multimodal large language models (MLLMs) to support individuals with disabilities in perceiving environmental risks, reasoning about potential hazards, and planning safe actions in everyday settings. This chapter presents the system’s overall architecture, workflow, and core agent modules, along with its implementation framework.

\begin{figure*}[htbp]
\centering
\includegraphics[width=1.0\textwidth]{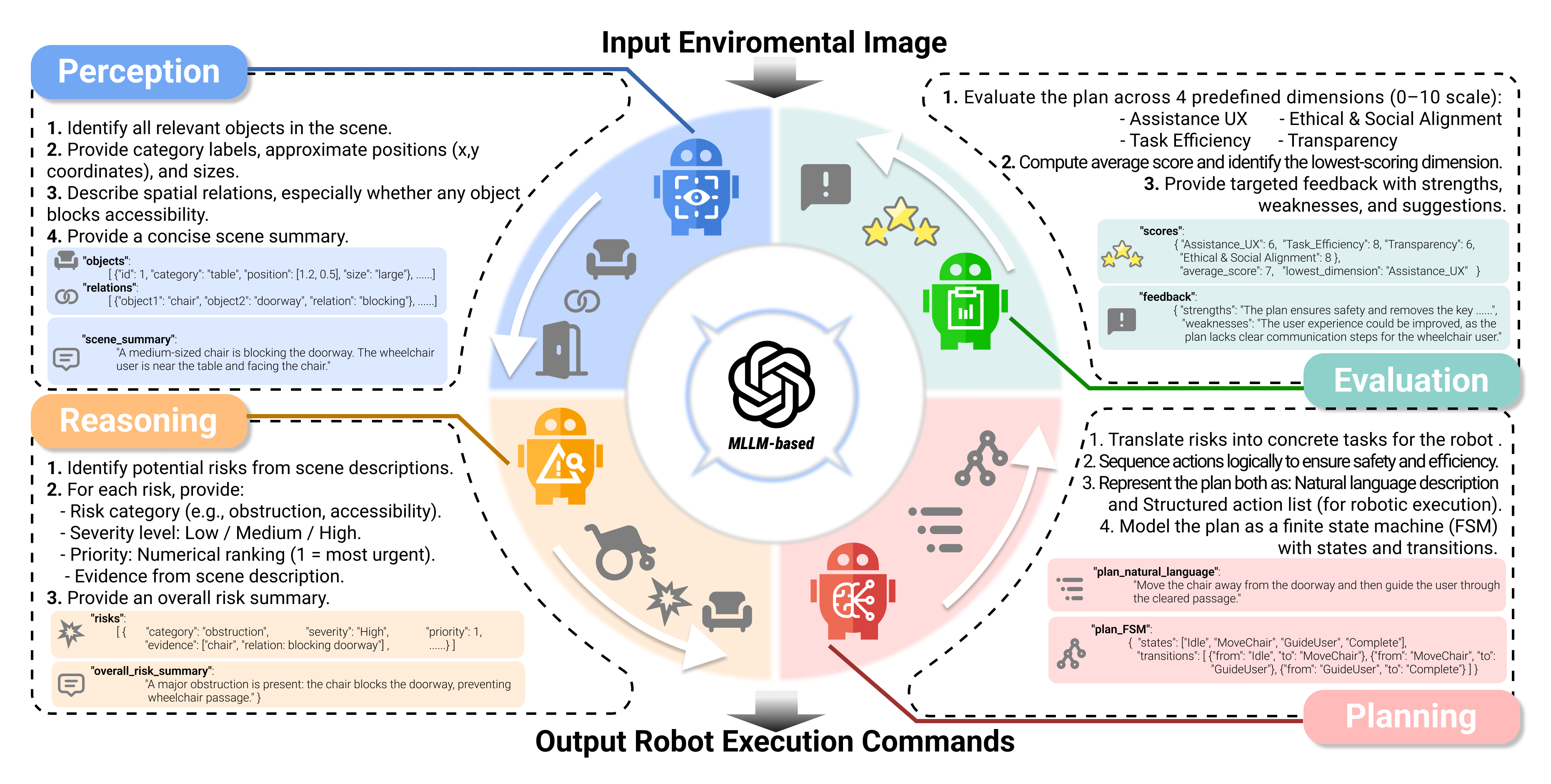}
\caption{Overview of the proposed system}
\label{figure:framework}
\end{figure*}

\subsection{Overall System Architecture}
The architecture follows a closed-loop cycle of perception, reasoning, planning, evaluation, iteration, ensuring coordination and continuous feedback among agents. As shown in Figure \ref{figure:framework}, the system is architected as a sequential pipeline comprising an input transduction layer, four core cognitive agent modules operating in a coordinated manner, and an output actuation interface.


The input layer is tasked with ingesting RGB images of the ambient environment, which can be optionally augmented with depth maps or precomputed semantic segmentation masks to improve the understanding of the geometric and semantic scenes of the system. The visual perception agent detects obstacles, furniture, users, and assistive devices, and evaluates pathway accessibility. Based on these outputs, the reasoning agent identifies risks and assigns severity and priority levels. The planning agent then generates executable action plans for robots, including obstacle removal, object adjustment, and path guidance. The feedback agent evaluates plan feasibility and safety, returning to the planning stage for revision when needed. Finally, the output interface delivers execution commands to the robot and provides alerts or guidance to the user or caregiver. This closed-loop design integrates perception, reasoning, planning, and feedback into a unified process, enhancing system robustness in dynamic and complex environments.

\subsection{Workflow}
The system workflow consists of four stages: perception, risk analysis, planning, and iterative optimization, as shown in Algorithm \ref{alg:system}. In the perception stage, the visual agent processes environmental images with multimodal models and detection algorithms to produce structured scene representations containing object categories, coordinates, and spatial relations. In the risk analysis stage, the reasoning agent detects hazards, such as blocked pathways, oversized furniture, or narrow passages, and prioritizes them by urgency and severity. The planning stage then generates preliminary action sequences or finite state machines executable by robots, covering tasks such as obstacle clearing, furniture repositioning, and safe navigation.


The planning evaluation loop constitutes a generate and test optimization algorithm. The Planning Agent acts as a proposal generator, while the Evaluation Agent implements a multi-dimensional scoring function (defined by ${S_1, S_2, S_3, S_4}$). The loop’s termination condition ($\bar{S}$ > threshold) and the heuristic feedback rule (suggestions based on $d_{min}$) together form a lightweight, heuristic search procedure that iteratively refines the action sequence $\mathcal{A}$ towards $\mathcal{A}^*$, ensuring Pareto like improvements across competing objectives. The final output includes both robot executable commands and user oriented feedback, ensuring operational safety while maintaining transparency in human–robot interaction.

\begin{algorithm}[!t]
\caption{Proposed Multi-Agent System Workflow}
\label{alg:system}
\textbf{Input}: Input image $I_{RGB}$ \\
\textbf{Output}: Optimized action sequence $\mathcal{A}^*$

\begin{algorithmic}[1]
\STATE \textbf{Initialize}: $t=0$, set termination flag = False.

\STATE \textbf{Visual Perception Agent}:
    \STATE Extract global semantic features $F_{CLIP}$ from $I_{RGB}$.
    \STATE Perform object segmentation to obtain $\{m_i, S_i, (x_i, y_i), B_i, conf_i\}$ in Eq.\ref{eq:3}.
    \STATE Integrate features into multimodal representation $\mathcal{X}_{Agent1}$.
    \STATE Generate structured scene description $\mathcal{D}_{Agent1}$.

\STATE \textbf{Risk Assessment and Reasoning Agent}:
    \STATE Receive $\{I_{RGB}, \mathcal{D}_{Agent1}\}$ as input.
    \STATE Identify potential risks (e.g., narrow passage, blocked passage, over-height objects).
    \STATE Compute severity, urgency, and overall risk scores.
    \STATE Rank risks and produce risk report $\mathcal{R} = \{r_1, r_2, ..., r_m\}$.

\WHILE{termination flag = False}
    \STATE \textbf{Planning Agent}:
        \STATE Translate each risk $r_k \in \mathcal{R}$ into task $T_k$.
        \STATE Map tasks into candidate actions from base library $\mathcal{A}_{base}$.
        \STATE Check feasibility of actions and compute priorities.
        \STATE Generate preliminary action sequence $\mathcal{A}$.

    \STATE \textbf{Evaluation and Optimization Agent}:
        \STATE Evaluate $\mathcal{A}$ along four dimensions $\{S_1, S_2, S_3, S_4\}$.
        \STATE Compute overall mean score $\bar{S}$ and lowest-scoring dimension $d_{min}$.
        \IF{$\bar{S}$ satisfies predefined acceptance threshold}
            \STATE Set $\mathcal{A}^* = \mathcal{A}$.
            \STATE Set termination flag = True.
        \ELSE
            \STATE Generate improvement suggestions based on $d_{min}$.
            \STATE Send feedback to Planning Agent for adjustment.
        \ENDIF
\ENDWHILE

\STATE \textbf{return} Optimized action sequence $\mathcal{A}^*$.
\end{algorithmic}
\end{algorithm}

\subsection{Multi-Agent Collaboration Module Design}
\subsubsection{Visual Perception Agent}
The visual perception agent is responsible for generating structured representations of the environment. It integrates a CLIP-based semantic extractor, a DeepV-based visual feature extractor, and the perception agent itself. Let the input RGB image be denoted as $I_{\text{RGB}} \in \mathbb{R}^{H \times W \times 3}$, where $H$ and $W$ represent the image height and width in pixels. At this stage, $I_{\text{RGB}}$ is simultaneously processed by two feature extraction modules, CLIP and DeepV3, which analyze the image from complementary perspectives.

\textbf{(1) CLIP module: Global semantic feature extraction.}
The CLIP module encodes the input image $I_{\text{RGB}}$ into a global semantic feature vector $F_{\text{CLIP}}$. Using a pretrained visual encoder, CLIP produces a high-dimensional embedding $F_{\text{CLIP}} \in \mathbb{R}^{D_{\text{CLIP}}}$, where $D_{\text{CLIP}} = 512$. This embedding captures semantic attributes such as scene type (e.g., indoor vs. outdoor) and the distribution of salient elements. The primary objective is to provide global semantic cues that support higher-level reasoning, expressed as:
\begin{equation}
F_{\text{CLIP}} = \text{CLIP}_{\text{encoder}}(I_{\text{RGB}})
\label{eq:1}
\end{equation}

\textbf{(2) DeepLV3 module: Instance segmentation and object detection.}
Complementing CLIP’s global features, the SAM module focuses on instance-level segmentation to delineate object boundaries at the pixel level. For each detected object $i$, the module outputs a binary segmentation mask $m_i \in \{0,1\}^{H \times W}$, from which the object pixel set $S_i = \{(x,y) \mid m_i(x,y) = 1\}$ is derived. The centroid of the object is then computed as
\begin{equation}
    (x_i, y_i) = \left( \frac{1}{|S_i|} \sum_{(x,y) \in S_i} x, \frac{1}{|S_i|} \sum_{(x,y) \in S_i} y \right)
\label{eq:2}
\end{equation}
where $|S_i|$ denotes the size of the pixel set. Additional features include the bounding box parameters $B_i = [x_{\text{min},i}, y_{\text{min},i}, x_{\text{max},i}, y_{\text{max},i}]$, which define the spatial extent of the object, and the segmentation confidence score $\text{conf}_i \in [0, 1]$, which quantifies the reliability of the segmentation.

\begin{figure}[htbp]
\centering
\includegraphics[width=0.8\textwidth]{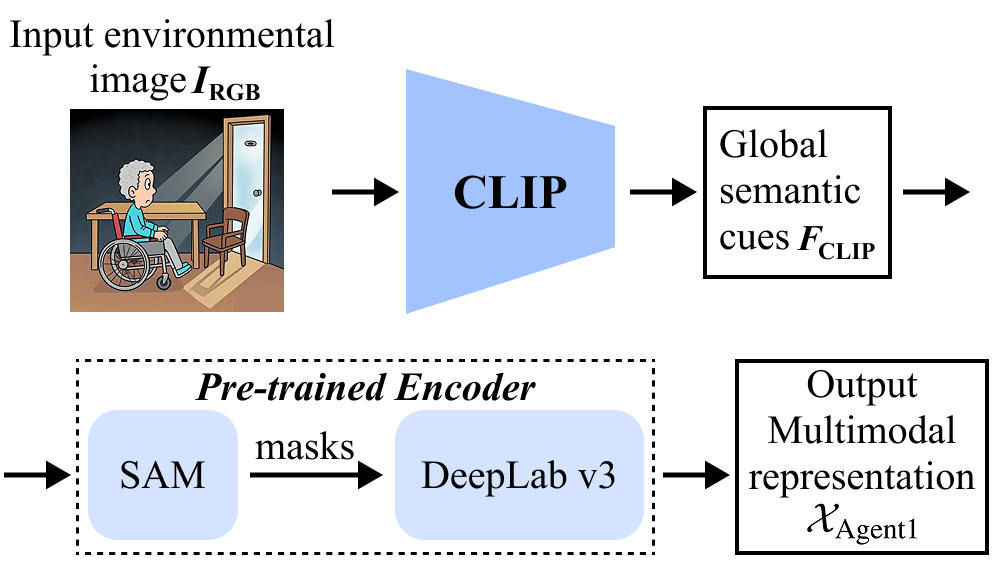}
\caption{Feature extraction and fusion framework}
\label{figure:encoder}
\end{figure}

\textbf{(3) Data integration.}
The outputs of CLIP and SAM are then integrated into a multimodal representation for Agent1:
\begin{equation}
    \mathcal{X}_{\text{Agent1}} = \{ I_{\text{RGB}}, F_{\text{CLIP}}, \{ m_i, S_i, (x_i, y_i), B_i, \text{conf}_i \}_{i=1}^n \}
\label{eq:3}
\end{equation}
where $n$ is the number of detected objects, see Figure \ref{figure:encoder} for details. Within this representation, $I_{\text{RGB}}$ provides the complete visual context, CLIP contributes global semantics, and SAM supplies object-level segmentation. By combining high-level semantic cues with detailed object boundaries, the design reduces the complexity of directly processing raw pixels and allows Agent1 to focus on semantic reasoning.

The representation $\mathcal{X}_{\text{Agent1}}$ is then processed by the MLLM, which leverages CLIP outputs to interpret SAM’s segmented objects. The result, denoted $\mathcal{D}_{\text{Agent1}}$, constitutes the Branch1 structured scene description. This includes an object set $\mathcal{O} = \{o_1, o_2, ..., o_n\}$, where each object $o_i$ is associated with category $c_i$ centroid $(x_i, y_i)$ , bounding box $B_i$, and size attributes; a channel attribute set $\mathcal{C} = \{width_c, direction_c, obstacle_c\}$ where $width_c$ denotes passage width and $obstacle_c$ lists channel obstructions; and user state information $\mathcal{U}$.

\subsubsection{Risk Assessment and Reasoning Agent}
\begin{longtable}{%
  >{\raggedright\arraybackslash}m{.1786\RiskW} 
  >{\centering\arraybackslash}m{.1786\RiskW}   
  >{\raggedright\arraybackslash}m{.5357\RiskW} 
  >{\centering\arraybackslash}m{.1071\RiskW}   
}
\caption{Risk Classification Table}\label{tab:risk}\\
\hline
\textbf{\small \makecell[c]{Risk\\Category}} &
\textbf{\small \makecell[c]{Severity\\Level}} &
\multicolumn{1}{>{\centering\arraybackslash}m{.5357\RiskW}}{\textbf{\small Case Description (Typical Scene)}} &
\multicolumn{1}{>{\centering\arraybackslash}m{.1071\RiskW}}{\textbf{\small Priority}} \\
\hline
\endfirsthead

\multicolumn{4}{c}{\tablename\ \thetable\ (continued)}\\
\hline
\textbf{\small \makecell[c]{Risk\\Category}} &
\textbf{\small \makecell[c]{Severity\\Level}} &
\multicolumn{1}{>{\centering\arraybackslash}m{.5357\RiskW}}{\textbf{\small Case Description (Typical Scene)}} &
\multicolumn{1}{>{\centering\arraybackslash}m{.1071\RiskW}}{\textbf{\small Priority}} \\
\hline
\endhead

\hline
\multicolumn{4}{r}{\small\itshape Continued on next page}\\
\endfoot

\hline
\endlastfoot

\multirow{3}{.1786\RiskW}{\raggedright Obstruction}
  & Low    & Object slightly encroaches the walkway; wheelchair can still pass without maneuver. & 3 \\
  & Medium & A chair narrows the passage; an alternative route exists but is less convenient.    & 2 \\
  & High   & The chair fully blocks the main doorway with no alternative path available.         & 1 \\ \hline

\multirow{3}{.1786\RiskW}{\raggedright Accessibility}
  & Low    & Target object is reachable with minor stretching/position adjustment.               & 3 \\
  & Medium & Target object is difficult to reach; assistance or careful repositioning is needed.& 2 \\
  & High   & The user cannot touch or operate a needed object (e.g., medication on a high shelf, switch too high). & 1 \\ \hline

\multirow{3}{.1786\RiskW}{\raggedright Collision Risk}
  & Low    & A static object is near but outside the current trajectory.                         & 3 \\
  & Medium & A moving obstacle (e.g., pet, person) is close to the planned path; avoidance is required. & 2 \\
  & High   & An object/person lies directly on the trajectory or within the immediate safety buffer.     & 1 \\ \hline

\multirow{3}{.1786\RiskW}{\raggedright User Safety Hazard}
  & Low    & Minor clutter; no likely harm if current behavior continues.                        & 3 \\
  & Medium & Potentially harmful element nearby (e.g., open drawer at knee level).               & 2 \\
  & High   & Immediate harm likely (e.g., wet/slippery floor on route, sharp edge protruding near user, unstable heavy object likely to fall). & 1 \\ \hline

\multirow{3}{.1786\RiskW}{\raggedright Navigation Feasibility}
  & Low    & Straightforward route; only small adjustments needed.                               & 3 \\
  & Medium & Complex maneuvers are needed but feasible (tight turns, temporary backing).         & 2 \\
  & High   & No feasible route without intervention or layout change.                             & 1 \\ \hline

\multirow{1}{.1786\RiskW}{\raggedright Others}
  & High   & Cases outside the predefined scope require human intervention to be requested.      & 1 \\ \hline
\end{longtable}

After obtaining structured perception results, the reasoning agent identifies and ranks potential safety risks such as blocked passages, insufficient space, or abnormal object heights. Risks are assessed by combining predefined rules (e.g., width thresholds, wheelchair standards) with MLLM-based semantic reasoning. This hybrid approach ensures both comprehensive coverage and contextual soundness. To maintain consistency, a predefined risk classification table (Table~\ref{tab:risk}) specifies typical categories, severity interpretations, and priorities.

The input of the reasoning agent is formally defined as
\begin{equation}
    \mathcal{X}_{\text{Agent2}} = \{I_{\text{RGB}}, \mathcal{D}_{\text{Agent1}}\}
\label{eq:4}
\end{equation}
where $I_{\text{RGB}} \in \mathbb{R}^{H \times W \times 3}$ is the original environment image, and $ \mathcal{D}_{\text{Agent1}}$ denotes the structured scene description output from \textit{Visual Perception Agent}.


The risk identification phase operates as a pattern matching and instantiation process. The agent performs a subgraph isomorphism check between the structured scene graph implicit in $\mathcal{D}{\text{Agent1}}$ and a library of predefined risk schema templates. Successful matches instantiate specific risk instances, which are then spatially localized within the coordinate frame of $I{\text{RGB}}$. Referring to the typical categories in (Table~\ref{tab:risk}), each scene element is systematically checked. For passage-related risks, the agent analyzes passage width and obstacle distribution against safety thresholds. Let the minimum wheelchair-accessible passage width be $\tau_w  (e.g., \tau_w = 0.8 \, \text{m})$. If the actual passage width satisfies $width_c < \tau_w$, the case is flagged as a “narrow passage risk.” If the obstacle set in the passage is nonempty, i.e., $obstacle_c \neq \emptyset$, the obstacle ratio $r_{obs}$ is computed as 
\begin{equation}
    r_{\text{obs}} = \frac{\sum_{o_i \in obstacle_c} \text{Area}(B_i)}{\text{Area}(\text{passage})} 
\label{eq:5}
\end{equation}
where $\text{Area}(B_i)$ denotes the bounding box area of obstacle $o_i$, and $\text{Area}(\text{passage})$ is the total passage area. When $r_{\text{obs}} > \tau_r$ (e.g., $\tau_r = 0.3$), the case is flagged as a “blocked passage risk.” For abnormal object height risks, the decision depends on the match between object vertical size and user interaction needs. Let the safety height threshold be $\tau_h$ (e.g., $\tau_h = 1.8 \, \text{m}$ for a standing user). If an object height $h_i > \tau_h$ and the object lies within the user’s activity region, it is flagged as an “over-height object risk.”

Risk levels are determined by combining the severity and urgency of identified risks according to the definitions in Table~\ref{tab:risk}. For the $k$-th risk, the severity score $s_{\text{sev},k}$ is assigned based on potential impact on user safety (e.g., complete passage blockage yields $s_{\text{sev},k} = 0.9$, while minor narrowing yields $s_{\text{sev},k} = 0.3$). The urgency score $s_{\text{urg},k}$ is derived from the spatiotemporal proximity between the user and the risk. If the user’s position is $(x_u, y_u)$ and the Euclidean distance to the risk region is $d_{u,k}$, then
\begin{equation}
    s_{\text{urg},k} = \exp\left( - \alpha \cdot d_{u,k} \right) 
\label{eq:6}
\end{equation}
where $\alpha$ is the distance decay coefficient (e.g., $\alpha = 0.5\, \text{m}^{-1}$). Closer distances yield higher urgency scores. The overall risk score $S_k$ is then computed as
\begin{equation}
    S_k = \omega_{\text{sev}} \cdot s_{\text{sev},k} + \omega_{\text{urg}} \cdot s_{\text{urg},k} 
\label{eq:7}
\end{equation}
where $\omega_{\text{sev}}$ and $\omega_{\text{urg}}$ are weighting factors satisfying $\omega_{\text{sev}} + \omega_{\text{urg}} = 1$ (e.g., 0.6 and 0.4, respectively). Based on the interval in which $S_k$ falls, the risk is assigned a level (e.g., $L_k = \text{High}$ for $S_k \geq 0.7$, and $L_k = \text{Medium}$ for $0.3 \leq S_k < 0.7$).

Prioritization is then determined by considering both risk level and impact scope. The priority index $P_k$ is defined as
\begin{equation}
    P_k = L_k \cdot \beta \cdot N_k 
\label{eq:8}
\end{equation}
where $L_k$ is the quantized risk level (high = 3, medium = 2, low = 1), $N_k$ is the number of users potentially affected (e.g., $N_k = 1$ in single-user scenarios), and $\beta$ is the scope coefficient (e.g., $\beta = 1.2$ for passage risks, $\beta = 0.8$ for local object risks). Risks are ranked in descending order of $P_k$, ensuring that highly severe and broadly impactful risks are prioritized for subsequent planning.

The final output of the reasoning agent is a structured risk report
\begin{equation}
    \mathcal{R} = \{r_1, r_2, ..., r_m\}
\label{eq:9}
\end{equation}
where each risk item is represented as $r_k = \{t_k, L_k, P_k, loc_k, desc_k\}$. Here, $t_k$ denotes the risk type (as defined in Table 1), $L_k$ the risk level, $P_k$ the priority index, $loc_k$ the spatial location, and $desc_k$ a natural language description (e.g., “An obstacle with width 0.5m is located at the center of the passage, severely obstructing movement”).

\subsubsection{Planning Agent}
The Planning and Task Allocation Agent serves as the core module for action generation, transforming the risk report into an executable structured action sequence. Its inputs are the risk report $\mathcal{R} = \{r_1, r_2, ..., r_m\}$ from the reasoning agent and the structured scene description $\mathcal{D}_{\text{Agent1}}$, while its output is an action sequence $\mathcal{A} = [a_1, a_2, ..., a_k]$ comprising action types, parameters, and execution order. For each risk $r_k$, a corresponding core task $T_k$ is generated, forming the task set $\mathcal{T} = \{T_k \mid T_k \leftrightarrow r_k, k=1,2,...,m\}$, e.g., a passage blockage risk $r_{obs}$ maps to task $T_{obs}=$ "Clear passage obstacles". 

Actions are selected from the base action library $\mathcal{A}_{\text{base}} = \{a_{\text{move}}, a_{\text{lift}}, a_{\text{guide}},...\}$, with each action $a_i$ parameterized as $a_i = \{t_i, \text{obj}_i, \text{pos}_i, \text{constraint}_i\}$, where $t_i$ denotes the action type, $\text{obj}_i$ the target object, $\text{pos}_i$ the goal position, and $\text{constraint}_i$ execution constraints (e.g., force limits). Feasibility is ensured via
\begin{equation}
    \text{Feasible}(a_i) = \begin{cases}1, & \text{obj}_i \in \mathcal{O} \cap \text{RobotCapable}(a_i) \\ 0, & \text{otherwise}\end{cases}
\label{eq:10}
\end{equation}
where $\text{RobotCapable}(a_i)$ indicates whether the robot can perform $a_i$.

Action priority is computed by weighting risk priority and execution cost. For action $a_i$ associated with risk priority $P_k$ and execution cost $C_i$ (time and energy), the priority index is
\begin{equation}
    \text{Pri}(a_i) = \omega_p \cdot P_k + \omega_c \cdot C_i
\label{eq:11}
\end{equation}
with $\omega_p + \omega_c = 1$. To maintain logical sequence integrity, state transitions follow $s_{i+1} = \delta(s_i, a_i)$, where $s_i$ is the current state and $\delta$ the state transition function. For example, a “move furniture → clear obstacles” transition requires completion of the prior state $s_i =$ “furniture in place”. The final action sequence $\mathcal{A}$ satisfies temporal constraints $\text{Order}(a_i) < \text{Order}(a_j) \Leftrightarrow i < j$, with execution parameters such as movement distance $d_{i} = \|\text{pos}_i - \text{pos}_{\text{current}}\|$ specified for each action.
\subsubsection{Evaluation and Optimization Agent}
The evaluation and optimization agent is responsible for assessing the quality of planning schemes and guiding their refinement, thereby improving overall performance through multidimensional evaluation and a closed-loop feedback mechanism. Departing from conventional unimodal feasibility verification, this agent implements a multi-criteria decision analysis (MCDA) framework predicated on four orthogonal evaluation dimensions. This framework facilitates a Pareto-optimal trade-off analysis among competing objectives, such as user-centric experience metrics and system-level robustness guarantees, during the holistic assessment of candidate plans. Its input is the preliminary action sequence $\mathcal{A} = [a_1, a_2, ..., a_k]$ generated by the planning and task allocation agent. The evaluation is conducted along four dimensions defined in Table 2, each associated with specific criteria (e.g., Assistance UX covers clarity of interaction, usability, and trust; Task Efficiency addresses time efficiency, resource utilization, and robustness), forming a multi-perspective assessment network.

At the core of the evaluation lies the quantitative scoring mechanism. Let the score for the $d$-th dimension be $S_d$, where $d=1,2,3,4$ correspond to the four dimensions and $S_d \in [0,10]$ (0 = worst, 10 = best). Scores are derived from semantic matching and rule-based mapping against the defined criteria (e.g., “clear conveyance of action intent” yields a high score under Assistance UX). The score vector is represented as $\mathbf{S} = [S_1, S_2, S_3, S_4]$. From these scores, two key indicators are computed: the overall mean score $\bar{S}$ and the lowest-scoring dimension $d_{min}$. The overall mean score is given by
\begin{equation}
    \bar{S} = \frac{1}{4} \sum_{d=1}^{4} S_d
\label{eq:12}
\end{equation}
 while the lowest-scoring dimension is determined by $d_{\text{min}} = \arg\min_{d \in \{1,2,3,4\}} S_d$. For example, if $\mathbf{S} = [8, 9, 7, 6]$, then $\bar{S} = 7.5$ and $d_{min}=4$, indicating that Ethical and Social Alignment is the weakest dimension.

The transformation of evaluation results into improvement suggestions must consider dimension-specific characteristics. For the lowest-scoring dimension $d_{min}$, targeted optimization directions are generated. The priority of each suggestion $P_{imp}$ is inversely proportional to the corresponding score:
\begin{equation}
    P_{\text{imp}}(d) = \frac{10 - S_d}{\sum_{d=1}^{4} (10 - S_d)}
\label{13}
\end{equation}

Suggestions with higher priority are fed back first to the planning and task allocation agent, guiding the next round of plan adjustments (e.g., optimizing interaction flow or simplifying action steps). The agent outputs an evaluation report, which includes the score vector $S$, overall mean score $\bar{S}$, lowest-scoring dimension $d_{min}$, and a list of improvement recommendations.

\section{Evaluation}
\subsection{Settings}
All experiments were conducted on a Windows workstation equipped with an Nvidia 4090 GPU, Intel Core i9-14900HX CPU, and 128 GB memory, where we deployed our framework and baselines (LLava-1.5, Qwen2-VL-7B, Qwen2-VL-72B, Qwen2.5-VL, Qwen3-235B-A, Llama-4-Scout, Llama-Vision, Llama-3.2-11B, Llama-3.2-90B, GPT-oss, GPT-4o). For comparative and ablation studies, we constructed a dataset of 800 indoor simple-scene images (Level 1), 800 indoor complex-scene images (Level 2), and 800 manga-style images (Level 3) generated via style transfer. Evaluation followed the strategy of Wang et al. \cite{LLM-SAP}, where GPT-5 and two human experts independently scored and ranked model outputs. For generalization experiments, we further tested on the dataset of Wang et al. (hereinafter denoted as LLM-SAP), Hazards\&Robots \cite{mantegazza2022outlier}, Home Fire Dataset \cite{homefire}, and our dataset, using a 7-point scale (1–7, higher is better) assessed by both GPT-5 and human experts.
·

\begin{table*}[ht]
\caption{Average Rankings of Different Models}
\label{tab:1}
\centering
\footnotesize  
\begin{tabular}{p{2.5cm}c*{8}{>{\centering\arraybackslash}p{0.8cm}}} 
\hline
\multirow{2}{*}{Model} & \multirow{2}{*}{Param.} & \multicolumn{4}{c}{AI (GPT-5)} & \multicolumn{4}{c}{Experts (sampling)} \\
\cline{3-10}
& & Overall & L1 & L2 & L3 & Overall & L1 & L2 & L3 \\
\hline
LLaVA-1.5 & 7B & 8.03 & 8.4 & 8.9 & 6.8 & 6.33 & 7 & 6.4 & 5.6 \\ 
Llama-Vision & 11B & 5.23 & 5.4 & 4.8 & 5.5 & 6.3 & 7.2 & 5.5 & 6.2 \\ 
Llama-3.2-11B & 11B & 5.77 & 5.1 & 5.8 & 6.4 & 6.2 & 6.1 & 6.7 & 5.8 \\ 
Llama-3.2-90B & 90B & 6.77 & 6.8 & 6.1 & 7.4 & 5.9 & 4.8 & 6.1 & 6.8 \\ 
Llama-4-Scout & 17B & 5.77 & 5.9 & 6.9 & 4.5 & 6.4 & 6.89 & 5.9 & 6.4 \\ 
Qwen-VL & 7B & 6.18 & 5.56 & 6.78 & 6.2 & 7.6 & 7.9 & 7.9 & 7 \\ 
Qwen2-VL & 72B & 8.12 & 8.56 & 7.3 & 8.5 & 7.2 & 6 & 7.9 & 7.7 \\ 
Qwen2.5-VL & 72B & 8.43 & 7.89 & 8.2 & 9.2 & 8.33 & 8 & 8.8 & 8.2 \\ 
Qwen3-235B-A & 22B & \underline{2.77} & \underline{3.4} & \underline{2.6} & \underline{2.3} & \underline{3.77} & \underline{2.9} & \underline{4} & \underline{4.4} \\ 
GPT-oss & 20B & 6.8 & 6.4 & 6.7 & 7.3 & 6.37 & 7.22 & 5.3 & 6.6 \\ 
Ours & 1800B & \textbf{1.93} & \textbf{1.7} & \textbf{2.2} & \textbf{1.9} & \textbf{1.33} & \textbf{1.2} & \textbf{1.5} & \textbf{1.3} \\ 
\hline 
\end{tabular}
\end{table*}

\subsection{Comparative Experiment}
\begin{figure}[htbp]
\centering
\includegraphics[width=0.75\textwidth]{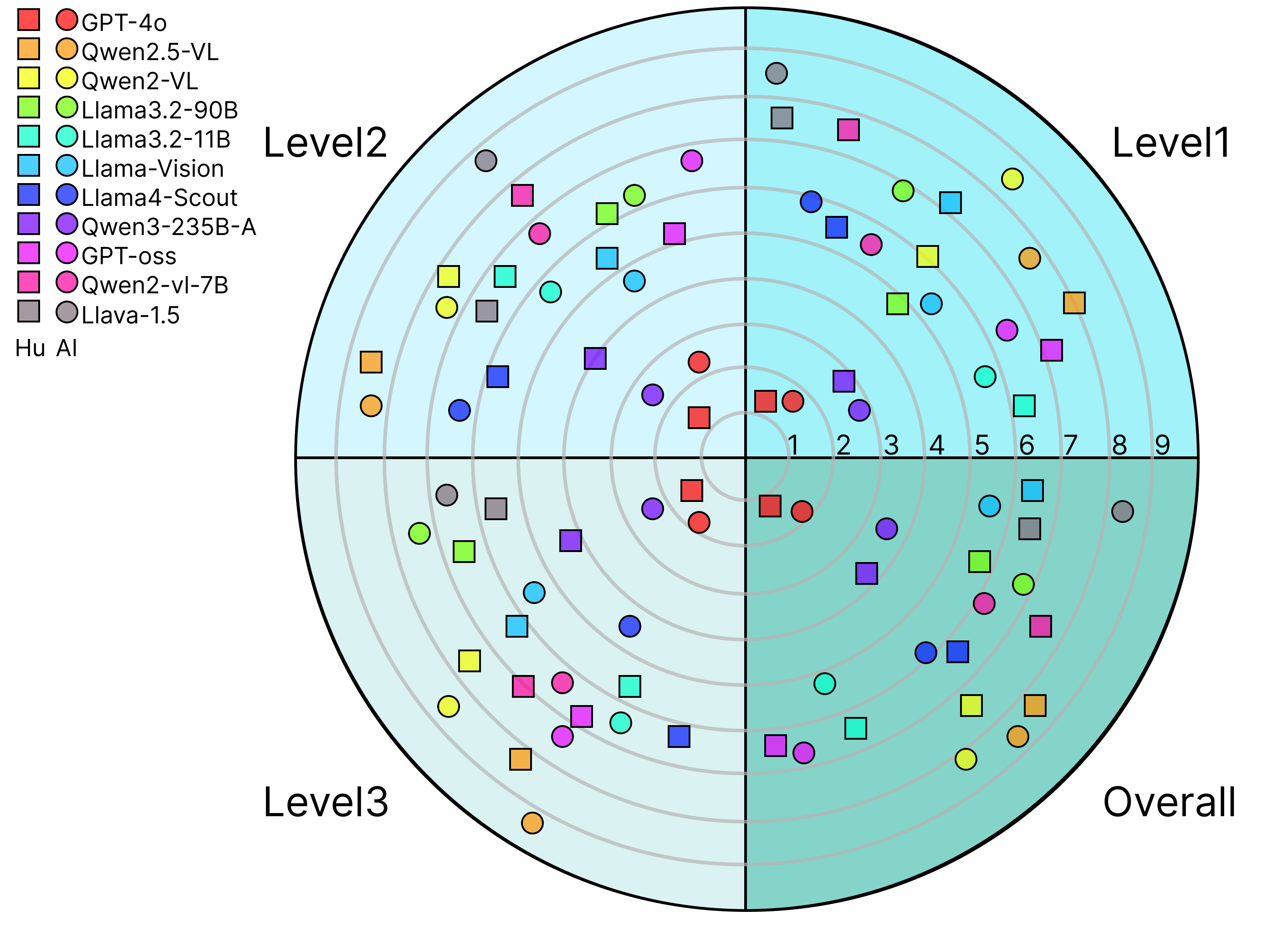}
\caption{Average ranking visualization of different models on four types of scenarios}
\label{Comp1}
\end{figure}

To identify the most suitable model for the constructed scenarios, we evaluated eleven mainstream large models, including LLaVA-1.5 (see Tab. \ref{tab:1} and Fig. \ref{Comp1} for details). The radar chart of the results is shown in Fig. \ref{Comp1}. GPT-4o achieved the best performance across all three scenarios under both AI and human expert evaluations. Its average rankings under the AI-based scoring (Overall, Level 1–3) were 1.93, 1.70, 2.20, and 1.90, while under the human expert scoring, they were 1.33, 1.20, 1.20, and 1.30, indicating that GPT-4o is the most suitable model for the target tasks in this domain.

\begin{figure}[htbp]
\centering
\includegraphics[width=0.85\textwidth]{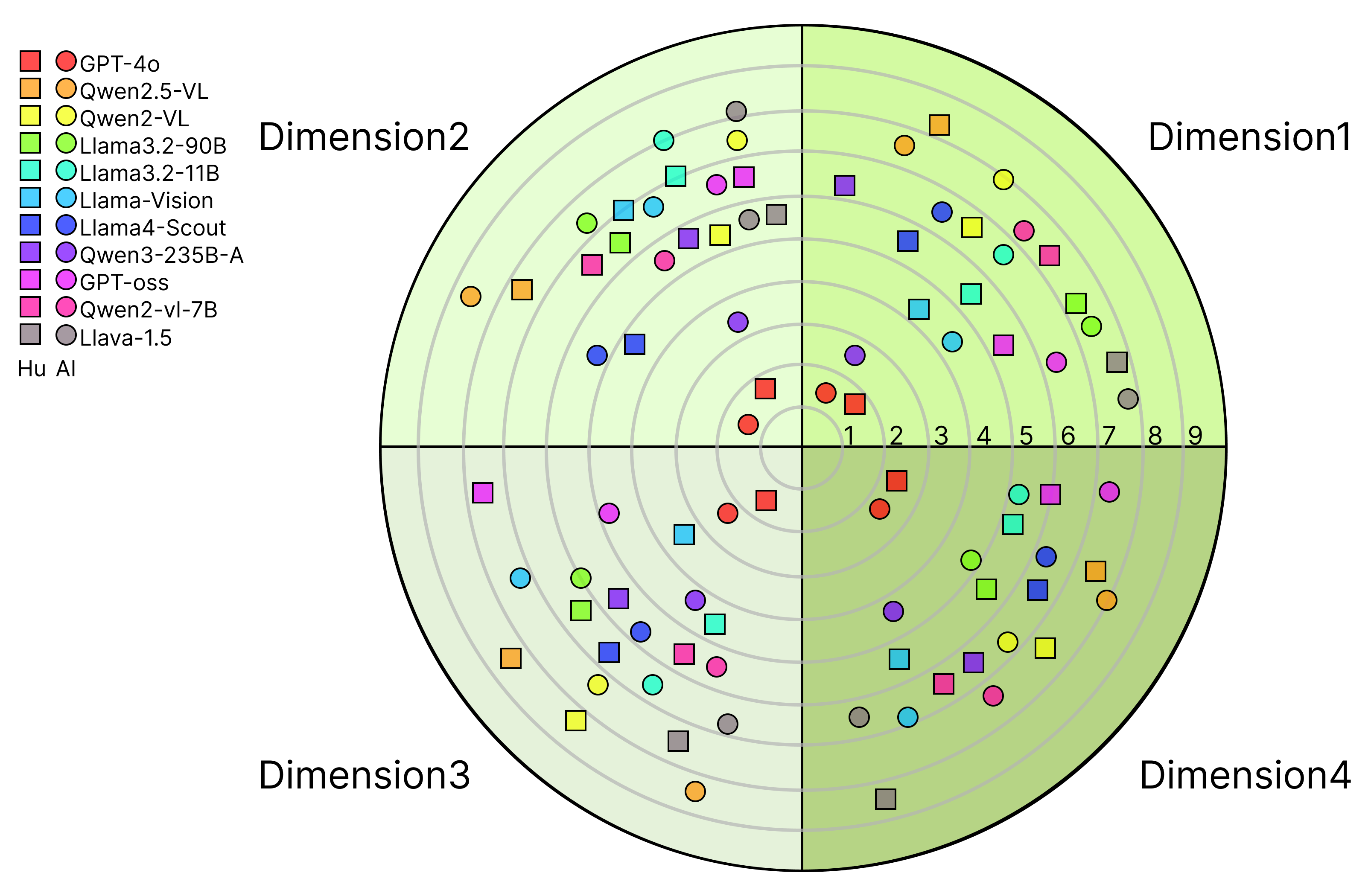}
\caption{Average ranking visualization of different models on four evaluation dimensions}
\label{Comp2}
\end{figure}

Moreover, we observed that the Qwen3-235B-A model consistently achieved the second-best average rankings across all scenarios. Its performance notably outpaced other models in the same series, such as Qwen2 and Qwen2.5, demonstrating the advantage of the updated architecture and parameter configuration. This trend is also visually evident in the radar chart in Fig. \ref{Comp1}, where the red markers representing GPT-4o’s performance are positioned closest to the chart’s center, followed by the purple markers for Qwen3-235B-A, indicating its stable and strong performance relative to the other evaluated models.

On the other hand, to further validate the performance of the proposed framework in information analysis and processing, we conducted comparative experiments against mainstream large models. Using both AI and human experts as evaluators, model outputs were ranked across four dimensions: Assistance UX, Task Efficiency, Transparency, and Ethical \& Social Alignment. The results, shown in Tab. \ref{tab:2} and Fig. \ref{Comp2}, demonstrate that our framework achieved the best rankings across all dimensions, with average scores of 1.89, 1.44, 2.22, and 2.11 under AI evaluation, and 1.67, 1.78, 1.44, and 2.10 under human expert evaluation, thereby confirming the effectiveness of the proposed framework.


\renewcommand{\arraystretch}{1.1}
\begin{table*}[ht]
\caption{Average rankings of Different Models on Four Evaluation Dimensions}
\label{tab:2}
\centering
\footnotesize
\begin{tabular}{p{1.5cm}c*{8}{>{\centering\arraybackslash}p{0.8cm}}} 
\hline
\multirow{2}{*}{Model} & \multirow{2}{*}{Param.} & \multicolumn{4}{c}{AI (GPT-5)} & \multicolumn{4}{c}{Experts (sampling)} \\
\cline{3-10}
& & D1 & D2 & D3 & D4 & D1 & D2 & D3 & D4 \\
\hline
LLaVA-1.5 & 7B & 7.89 & 5.56 & 6.89 & 6.78 & 7.67 & 5.89 & 7.22 & 8.2 \\ 
Llama-Vision & 11B & 4.33 & 6.78 & 7.11 & 6.89 & \underline{4.34} & \underline{4.89} & \underline{3.22} & 5.3 \\ 
Llama-3.2-11B & 11B & 6.67 & 8 & 6.78 & 5.22 & 5.11 & 7 & 4.56 & \underline{5.2} \\ 
Llama-3.2-90B & 90B & 7.33 & 7.22 & 6 & 4.78 & 7.23 & 6.32 & 6.11 & 5.5 \\ 
Llama-4-Scout & 17B & 6.56 & 5.22 & 5.78 & 6.13 & 5.78 & 5.89 & 6.78 & 6.3 \\ 
Qwen-VL & 7B & 7.25 & 5.44 & 5.44 & 7.22 & 7.22 & 7.67 & 5.67 & 6.3 \\ 
Qwen2-VL & 72B & 8 & 7.33 & 7.33 & 6.89 & 6.45 & 5.44 & 8.44 & 7.1 \\ 
Qwen2.5-VL & 72B & 7.56 & 8.67 & 8.33 & 8 & 8.23 & 7.89 & 8.67 & 7.9 \\ 
Qwen3-235B-A & 22B & \underline{2.67} & \underline{3.67} & \underline{4.22} & \underline{4.33} & 5.22 & 6.75 & 5.89 & 6.2 \\ 
GPT-oss & 20B & 6.33 & 6.67 & 4.78 & 7.11 & 6.45 & 6.75 & 7.89 & 6 \\ 
Ours & 1800B & \textbf{1.89} & \textbf{1.44} & \textbf{2.22} & \textbf{2.11} & \textbf{1.67} & \textbf{1.78} & \textbf{1.44} & \textbf{2.1} \\ 
\hline 
\end{tabular}
\vspace{-1.5em}
\end{table*}

\renewcommand{\arraystretch}{1.1}
\begin{table*}[htbp]
\caption{Ablation study results, indicating that our method achievs the best performance.}
\label{tab:3}
\centering
\footnotesize
\begin{tabular}{p{2.6cm}*{8}{>{\centering\arraybackslash}p{0.9cm}}} 
\hline
\multirow{2}{*}{Condition} & \multicolumn{4}{c}{AI (GPT-5)} & \multicolumn{4}{c}{Experts (sampling)} \\
\cline{2-9}
& Overall & L1 & L2 & L3 & Overall & L1 & L2 & L3 \\
\hline
No Branch 1 & 6.97 & 5 & 8 & 7.9 & 7.3 & 5.7 & 8.3 & 7.9 \\ 
No Branch 2 & 5.17 & \underline{4.8} & 4.7 & 6 & 5.17 & 4.4 & 5.1 & 6 \\ 
No Branch 3 & \underline{4.4} & 6.2 & \underline{3} & \underline{4} & 4.67 & 7 & \underline{3} & \underline{4} \\ 
No Branch 4 & 5.43 & 5.1 & 5.4 & 5.8 & \underline{4.03} & \underline{2.4} & 3.9 & 5.8 \\ 
No Branch 1\&2 & 5.93 & 6.5 & 6.5 & 4.8 & 5.97 & 6.8 & 6.3 & 4.8 \\ 
No Branch 1\&3 & 6.8 & 7.1 & 6.5 & 6.8 & 6.5 & 6.4 & 6.3 & 6.8 \\ 
No Branch 1\&4 & 6 & 6.5 & 6.3 & 5.2 & 6.5 & 7.1 & 7.2 & 5.2 \\ 
No Branch 2\&3 & 5.97 & 6 & 6.5 & 5.4 & 6.27 & 6.4 & 7 & 5.4 \\ 
No Branch 2\&4 & 8.07 & 6.9 & 8 & 9.3 & 8.5 & 7.9 & 8.3 & 9.3 \\ 
No Branch 3\&4 & 9.23 & 9.8 & 8.9 & 9 & 9.27 & 9.7 & 9.1 & 9 \\ 
Ours & \textbf{1.73} & \textbf{1.4} & \textbf{2} & \textbf{1.8} & \textbf{1.53} & \textbf{1.3} & \textbf{1.5} & \textbf{1.8} \\ 
\hline 
\end{tabular}
\vspace{-1.0em}
\end{table*}

\subsection{Ablation Experiment}
To validate the effectiveness and necessity of the designed multi-branch structure, we conducted ablation experiments by removing individual agents to assess model performance. The same evaluation procedures as in the previous sections were applied. As shown in the results Tab. \ref{tab:3}, the complete strategy (last row) consistently achieved the best performance under both AI and human expert rankings, with scores of 1.73, 1.40, 2.00, and 1.80 for AI evaluation, and 1.53, 1.30, 1.50, and 1.80 for human expert evaluation. Furthermore, by comparing the first four rows, we observe that removing any branch degrades performance, with the absence of Branch 1 causing the most significant drop. This highlights the critical role of this perception module, consistent with the theoretical expectation that Branch 1 provides the foundational input for all subsequent branches. These results indicate that the four-stage optimization strategy effectively supports precise environmental analysis and generates accurate action recommendations.


\subsection{Generalization Experiment}
To investigate the generalization capability of the proposed model, we conducted experiments on multiple cross-domain datasets. In the results Tab. \ref{tab:4}, it is notable that the proposed system consistently achieved scores above the neutral midpoint (4 on the 1–7 scale) across all evaluated datasets and scenario types, reflecting its robust generalization capability. In particular, for our own dataset, which includes three levels of increasing scene complexity, the framework obtained the highest AI-based score of 5.53 for Level 1 and a slightly lower score of 4.98 for Level 3, illustrating that even under visually and semantically challenging scenarios, the model maintains reliable decision-making. Expert evaluations further confirm this trend, assigning higher scores overall, especially for the more complex and stylized scenes, which may indicate that the system’s behavior decisions align well with human expectations in realistic and abstracted contexts.

Overall, these results indicate that the proposed framework generalizes across diverse environments, producing scenario-aware, human-aligned action recommendations robust to varying scene complexities. This adaptability supports various assistive and safety-critical tasks, from indoor navigation to emergency hazard mitigation, while aligned with human expectations and operational requirements.


\renewcommand{\arraystretch}{1.1}
\begin{table}[htbp] 
\caption{Generalization Performance of the System across Datasets (Scale: 1--7)}
\label{tab:4}  
\centering
\footnotesize
\begin{tabular}{p{2.5cm}p{5cm}c*{2}{>{\centering\arraybackslash}p{1.2cm}}} 
\hline
Dataset & Scenario & Size & \multicolumn{1}{c}{AI Score} & \multicolumn{1}{c}{Expert Score} \\
\hline
LLM-SAP & children's hazard scenarios & 540 & 5.03 & 4.89 \\ 
Hazards\&Robots & indoor anomaly scenarios & 10000 & 4.75 & 4.9 \\ 
Home fire & home fire scenarios & 1300 & 4.83 & 4.78 \\ 
\multirow{3}{*}{Ours} & indoor simple scenes (Level 1) & 800 & 5.53 & 5.88 \\ 
& indoor complex scenes (Level 2) & 800 & 5.18 & 6.28 \\ 
& manga-style scenes (Level 3) & 800 & 4.98 & 6.14 \\ 
\hline 
\end{tabular}
\vspace{-1.0em}
\end{table}

\section{Discussion and Conclusion}
In recent years, large language models have demonstrated high intelligence across multiple domains. However, single-modality models often fall short in complex scenarios, driving the emergence of multimodal large language models. MLLMs are capable of handling diverse and sophisticated tasks, showing significant promise in indoor assistive agent scenarios. Despite their potential, current research exhibits several limitations. Key challenges in this domain include accurate perception and dynamic risk assessment, efficient and robust multi-agent collaboration, and the lack of dedicated evaluation datasets. To address these issues, this work presents a MLLM-based multi-agent collaborative system that leverages a multi-branch architecture to perform environment information processing, risk assessment, and action planning. This framework offers a technical solution for domain-specific challenges and the needs of assistive applications for individuals with disabilities, while providing a foundation for future research in related areas.

\section*{CRediT authorship contribution statement}

\noindent\textbf{Renjun Gao:} Conceived the study; designed the multi-agent architecture; implemented system integration; developed the risk-assessment and evaluation modules; curated datasets and experimental protocols; ran the main experiments; drafted the manuscript; conducted ablation analyses; contributed to the writing and revisions; prepared the figures.

\section*{Funding Statement}
This research received no specific grant from any funding agency in the public, commercial, or not-for-profit sectors.

\section*{Declaration of Competing Interests}
The authors declare that there are no competing financial interests or personal relationships that could have influenced the work reported in this paper.




\bibliographystyle{elsarticle-harv}  
\bibliography{cas-refs}  

\end{document}